\documentclass[10pt,twocolumn,letterpaper]{article}

\pdfoutput=1
\usepackage{wacv}
\usepackage{times}
\usepackage{epsfig}
\usepackage{graphicx}
\usepackage{amsmath}
\usepackage{amssymb}

\usepackage{multicol}

\usepackage{subfiles}

\usepackage{enumitem}
\usepackage[export]{adjustbox}
\usepackage{dsfont}

\usepackage[accsupp]{axessibility}

\usepackage{algorithm}
\usepackage{algorithmic}

\usepackage{multirow}
\usepackage{booktabs} 

\iftrue
\newcommand{\alertJW}[1]{{\color{magenta}{\bf JW:#1}}}
\newcommand{\JZ}[1]{{\color{blue}{\bf JZ:#1}}}
\newcommand{\BR}[1]{{\color{red}{\bf Bog:#1}}}

\else
\newcommand{\alertJW}[1]{}
\newcommand{\JZ}[1]{}
\newcommand{\BR}[1]{}
\fi

\newcommand{\minisection}[1]{\vspace{0.03in} \noindent {\bf #1}}

\def\wacvPaperID{541} 

\wacvfinalcopy 

\ifwacvfinal
\usepackage[breaklinks=true,bookmarks=false]{hyperref}
\else
\usepackage[pagebackref=true,breaklinks=true,colorlinks,bookmarks=false]{hyperref}
\fi

\begin{document}

\title{Class-Balanced Active Learning for Image Classification}

\author{Javad Zolfaghari Bengar$^{1,2}$
\and Joost van de Weijer$^{1,2}$
\and Laura Lopez Fuentes$^{1}$
\and Bogdan Raducanu$^{1,2}$\\
Computer Vision Center (CVC)$^1$, Univ. Aut\`{o}noma of Barcelona (UAB)$^2$\\

{\tt\small \{jzolfaghari,joost,llopez,bogdan\}@cvc.uab.es}
}

\maketitle
\begin{abstract}
\vspace{-2mm}

Active learning aims to reduce the labeling effort that is required to train algorithms by learning an acquisition function selecting the most relevant data for which a label should be requested from a large unlabeled data pool. Active learning is generally studied on balanced datasets where an equal amount of images per class is available. However, real-world datasets suffer from severe imbalanced classes, the so called long-tail distribution. We argue that this further complicates the active learning process, since the imbalanced data pool can result in suboptimal classifiers. To address this problem in the context of active learning, we proposed a general optimization framework that explicitly takes class-balancing into account. Results on three datasets showed that the method is general (it can be combined with most existing active learning algorithms) and can be effectively applied to boost the performance of both informative and representative-based active learning methods. In addition, we showed that also on balanced datasets our method \footnote{Our code is available at: \url{https://github.com/Javadzb/Class-Balanced-AL.git}} generally results in a performance gain. 
\end{abstract}

\begin{figure*}[ht]
    \centering
    \includegraphics[width=.75\textwidth]{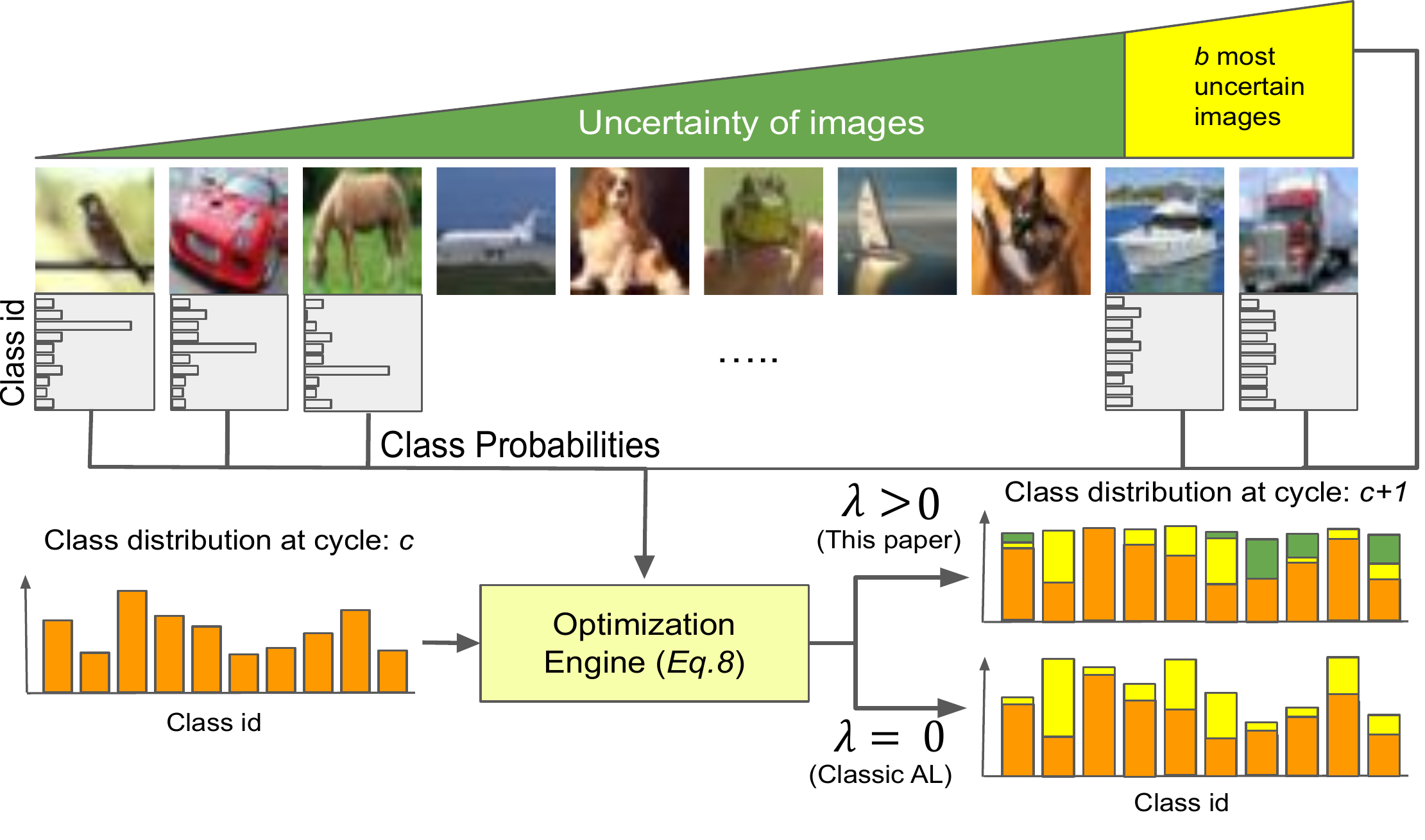}
    \vspace{-2mm}
    \caption{\small \textbf{Overview of our active learning framework.} The unlabeled samples are sorted by their uncertainty from green to yellow in ascending order. Given the the uncertainty of unlabeled samples and class distribution at cycle $c$, we propose to solve an optimization problem ($\lambda>0$) yielding samples that are simultaneously informative and form a balanced class distribution for training. Our sampling selects samples with lower uncertainty (in green) in addition to high uncertainty to improve class-balanced profile. In contrast, classical AL methods ($\lambda=0$) selects the most uncertain samples (in yellow) that result in an informative yet imbalanced training set.}
    \vspace{-4mm}
    \label{fig:AL_framework}
\end{figure*}
\vspace{-3mm}
\section{Introduction}
\vspace{-1mm}
Neural networks obtain state-of-the-art results on several computer vision tasks such as large-scale object detection \cite{Peng2020detection} or VQA \cite{Wang2020vqa}. 
However, the training of these  often very large networks requires large-scale labeled datasets, that are labor intensive and expensive to construct. Generally, in real-world the amount of data that could be labeled is literary unlimited (e.g. in autonomous driving, or robotics applications). Given an initial labeled dataset, deciding what new data to label from the unlabeled data pool is a relevant research question addressed by active learning. It aims to minimize the labeling effort while maximizing the obtained performance of the machine learning algorithm. Active learning has successfully been shown to reduce the labeling effort for image classification \cite{beluch2018ensembles,sener2018active}, object detection \cite{zolfaghari2019temporal}, regression \cite{denzler2018bmvc}, and semantic segmentation \cite{wang2020ss,kitani2020ss}.


Several query strategies have been proposed for sample selection. The most popular ones are those based on informativeness \cite{Yoo_2019_CVPR} and representativeness \cite{sener2018active} which demonstrated to be efficient for the task of selecting the most valuable samples. The informativeness criteria is responsible for selecting those samples which are the most uncertain (usually characterized by high-entropy) because they affect the generalization capability of the model (they are the ones which are mostly confusing the classifier, especially at the start of the active learning process when the number of labeled samples is small), while representativeness guarantees a diversity of the samples, following the underlying data distribution of the unlabeled data pool. 

Visual recognition datasets are often almost uniformly distributed (e.g. CIFAR \cite{krizhevsky2012learning} and ILSVRC \cite{krizhevsky2012imagenet}). However, for many real-world problems data follows a long-tail distribution \cite{newman2005power}, meaning that a small number of head-classes are much more common than a large number of tail-classes (e.g. iNaturalist~\cite{van2018inaturalist}, landmarks~\cite{noh2017large}). 
Classification on such imbalanced dataset is an important research topic~\cite{huang2016learning,cui2018large,ren2018learning}. However, active learning is mostly studied on curated close to uniform datasets. Given the predominance of long-tail distributions, especially for real-world applications in which active learning is a crucial capability, we here study active learning for imbalanced datasets. The aim is to minimize the labeling effort, while maximizing performance when measured on a balanced test set. 

Closely related to the class-imbalance dataset problem, is the sampling bias problem which is a well-documented drawback of active learning~\cite{mackay1992information,dasgupta2008hierarchical}. Datasets collected by active learning algorithms break the assumption that the data is identically and independently distributed (i.i.d), since the active learning algorithm might be biased towards particular regions of the unlabeled data manifold. One possible consequence of the sampling bias can be that the distribution over the classes does no longer follow that of the unlabeled data pool. Several papers have investigated this aspect of active learning however it remains not fully understood~\cite{beygelzimer2009importance,farquhar2021statistical}.
To mitigate the problems caused by the sampling bias and imbalanced datasets, in the current paper we introduce an optimization framework which corrects the class-imbalance presented in the unlabeled data pool, and aims to bias instead our selected samples to resemble the uniform distribution of the test set. The overview of the proposed approach is depicted in figure \ref{fig:AL_framework}. Since we have no access to the class labels of the unlabeled data, we propose to trust the predicted labels, and use them to select a set of class-balanced images. This combination leads to a minimization problem, which can be formalized as a binary programming problem. We show that our optimization scheme is efficient, boosting the performance of both informativeness and representativeness methods. 
In summary, the main contributions of this paper are:

\begin{itemize}[noitemsep,nolistsep]
\item We propose a novel active learning method for imbalanced unlabeled dataset that encourages the selection of class-balanced samples. 
\item The proposed optimization method is general and can be applied to both informativeness and representativeness based methods. 
\item Extensive experiments show that our method improves performance of active learning on imbalanced datasets. We show that even for balanced datasets the proposed method can lead to improvements, mostly by countering the sampling bias introduced by active learning. 
\end{itemize}
\begin{figure*}[ht]
    \centering
    \includegraphics[width=\textwidth]{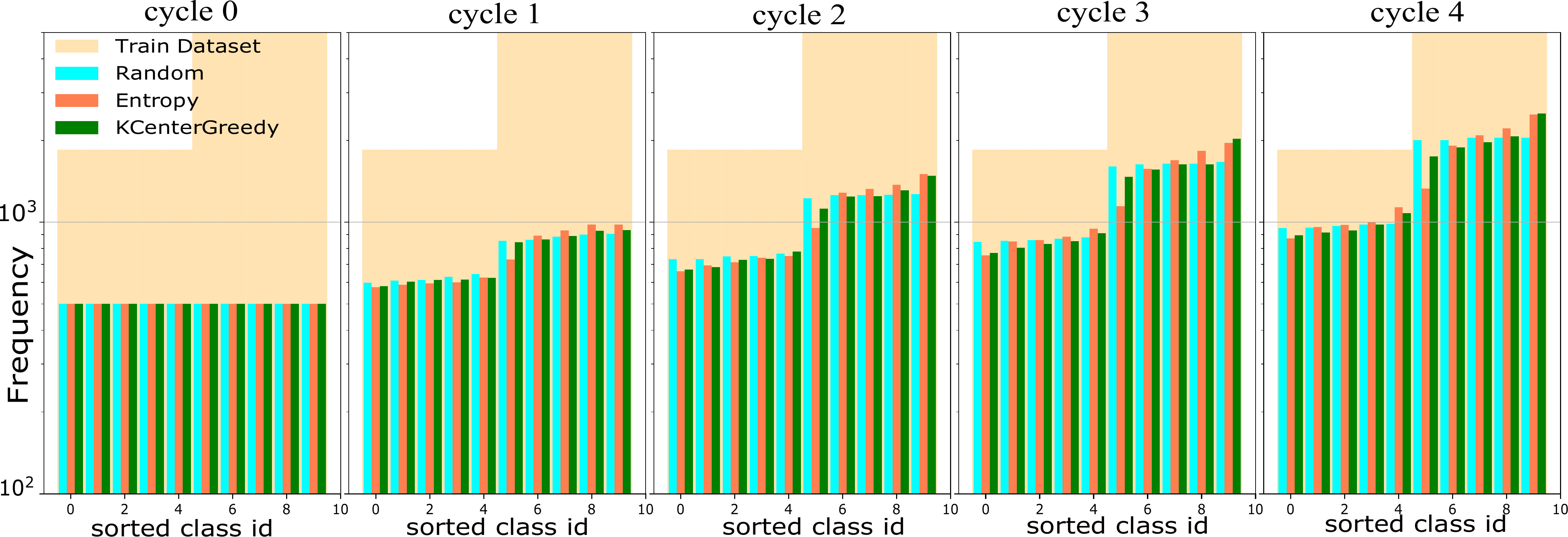}
    \caption{\small \textbf{Biased sampling across four AL cycles.} In the background, the imbalanced dataset is illustrated in yellow. The class distributions for two active learning approaches and random sampling are shown. Similar to Random sampling (in cyan), samples selected by active learning algorithms follow the biased distribution. Results are on imbalanced CIFAR10 (IF=0.3).}
    \label{fig:hist1}
    \vspace*{-2mm}
\end{figure*}
\vspace{-1mm}
\section{Related Work}\label{sec:related}
\vspace{-1mm}
\minisection{Active Learning.} 
Active Learning has been widely studied in various applications such as image classification \cite{urtasun2007al,snoek2015transfer,wu2018kdd} (including medical image classification \cite{falcao2015pr} and scene classification \cite{lin2014microsoft}), image retrieval \cite{zhang2010retrieval}, image captioning \cite{Deng2018al}, object detection \cite{zolfaghari2019temporal}, and regression \cite{freytag2014influential,denzler2018bmvc}. 
Strategies can be divided in three main categories: informativeness \cite{yang2015multi,gal2017icml,guo2010nips, gu2014modelchange,bengar2021deep}, representativeness \cite{falcao2015pr,sener2018active} and
hybrid approaches \cite{zhou2014hybrid,loog2018maxvariance}. A comprehensive survey of these frameworks can be found in \cite{settles2012active}.

Among all the aforementioned strategies, the informativeness-based approaches are the most successful ones, with uncertainty being the most used selection criteria used in both bayesian \cite{gal2017icml} and non-bayesian frameworks \cite{li2013adaptive}. In \cite{gal2017icml}, they obtain uncertainty estimates through multiple forward passes with Monte Carlo Dropout, but it is computationally inefficient for recent large-scale learning as it requires dense dropout layers that drastically slow down the convergence speed. More recently, \cite{ash2019deep} measures the uncertainty of the model by estimating the expected gradient length. On the other hand, \cite{Yoo_2019_CVPR,li2020learning} employ a loss module to learn the loss of a target model and select the images based on their output loss. 


Representativeness-based methods rely on selecting examples by increasing diversity in a given batch \cite{dutt2016active}. The Core-set technique  \cite{sener2018active} selects the samples by minimizing the Euclidian distance between the query data and labeled samples in the feature space. The Core-set technique is shown to be an effective method, however, its performance is limited by the number of classes in the dataset. Furthermore, it is less effective due to feature representation in high-dimensional spaces since p-norms suffer from the curse of dimensionality \cite{donoho2000high}. In a different direction, \cite{Sinha_2019_ICCV} uses an adversarial approach for diversity-based sample query, which samples the data points based on the discriminator's output, seen as a selection criteria. With the recent advancements in self-supervised learning, \cite{bengar2021reducing} integrated active learning with self-supervised pre-training. 

\minisection{Class-Imbalanced Data.} Learning with class-imbalanced data is a well investigated research problem~\cite{japkowicz2000class}. There are several approaches to address the conflict between a highly imbalanced training dataset and the objective to perform equally well for all classes on the test set. The bias towards the most frequent classes can be reduced by \emph{re-weighting} samples in the training objective. One popular approach is re-weighting samples by the inverse of their class-frequency~\cite{huang2016learning}. Cui et al.~\cite{cui2018large} improve upon this method, and propose to re-weight samples with the  effective number of its class. Another approach is based on \emph{re-sampling} where samples of rare classes are more often rehearsed during training~\cite{he2009learning}. Ren et al.~\cite{ren2018learning} investigate the training on imbalanced data in combination with label noise. They propose a method based on meta-learning that learns to assign weights to training examples. Our proposed method aims to prevent the dataset imbalance that could arise during the active learning cycles. 
We show that incorporating class-balance as one of the objectives of AL is of key importance on imbalanced datasets.

Previous works that addressed class imbalance in AL include \cite{wang2020important, aggarwal2020active,bhattacharya2019generic, yu2018active}. Among them, only \cite{aggarwal2020active} is applied to deep learning. Nevertheless, it studies sequential AL as balancing is performed during manual labeling making it practically infeasible for batch mode AL. In the same line ~\cite{c2018active} lacks automatic model to address the class-imbalance problem and the evaluations are human-centered only. Unlike \cite{choi2021vab} that lacks evaluation on large scale dataset, we show our method extends to Tiny ImageNet as a large dataset with diverse classes.

\section{Class Imbalance in Active Learning}\label{CBAL}
\subsection{Active Learning Setup}
Given a large pool of unlabeled data $\mathcal{D_U}$ and a total annotation budget $B$, the goal is to select $b$ samples in each cycle to be annotated to maximize the performance of a classification model. In general, AL methods proceed sequentially by splitting the budget in several \emph{cycles}.
Here we consider the batch-mode variant~\cite{settles2012active}, which annotates $b$ samples per cycle, since this is the only feasible option for CNN training.
At the beginning of each cycle, the model is trained on the labeled set of samples $\mathcal{D_L}$.
After training, the model is used to select a new set of samples to be annotated at the end of the cycle via an \emph{acquisition function}. 
The selected samples are added to the labeled set $\mathcal{D}_L$ for the next cycle and the process is repeated until the annotation budget $b$ is spent.
The acquisition function is the most crucial component and the main difference between AL methods in the literature. In the remainder of this section, we describe the motivation behind our proposed acquisition function. 
\vspace*{-2mm}
\subsection{Motivation}
Most active learning methods propose efficient sampling methods that are class agnostic. The underlying assumption is that the distribution of train and test datasets are uniform. However, in real world scenarios, where the datasets might be heavily imbalanced, the methods suffer from biased sampling towards the majority class. AL methods tend to sample more from frequent classes and less from minority classes which consequently leads to biased predictions and a performance drop. Fig.~\ref{fig:hist1} presents an example of a such dataset with various AL methods (see Suppl. Mat. \ref{hist_5_cycles:cifa100}). As it can be seen, the distribution of samples selected by both informative and representative based methods follow the distribution of the unlabeled dataset. Moreover the imbalance of selected samples grows across the cycles. It is known that when we aim for good performance on all classes these imbalanced training sets are suboptimal~\cite{huang2016learning,cui2018large}. We tackle the problem of class imbalance in the remainder of this section. 

\subsection{Reducing Class Imbalance} 
 A balanced set of samples requires an equal number of samples per class. Since we have no access to the class labels, we make an estimate of distribution of samples by using a probability matrix.
 Assume we have $|\mathcal{D_U}|=N$ unlabeled samples in $C$ categories. We use the classifier to output the softmax probability matrix $P$ on the unlabeled samples:
\begin{equation}
\label{eq:pairwise}
   P=
\begin{bmatrix}
    p_{11} & p_{12} & ... & p_{1C}\\
    p_{21} & p_{22} & ... & p_{2C}\\
    \vdots & \vdots & \ddots & \vdots\\ 
    p_{N1} & p_{N2} & ... & p_{NC}\\ 
\end{bmatrix}
\in R^{N\times C}
\end{equation}
where each row sums to 1. Similar to \cite{elhamifar2013convex}, we use variable $z_{i}\in \{0,1\}$ associated to sample $i$ to indicate whether a sample $i$ is selected or not. To measure the distance between the estimated distribution and the desired distribution we employ $\ell1$ norm as:
\begin{equation}
\label{eq:balterm}
\ell1(\Omega, P^T \boldsymbol{z})=\| \Omega(c)-P^T \boldsymbol{z} \|_{1}.
\end{equation}
Here $\Omega(c)$ is vector with components specifying the number of required samples from each class in order to achieve balance at cycle $c$. Given the labels of samples selected in previous cycles, it is straightforward to compute the samples required at cycle $c$: 
\begin{equation} \label{eq:omega}
\Omega(c) =[\omega_{1},\omega_{2}, ...,\omega_{C}],
\end{equation}
where,
\begin{equation} \label{eq:thresh}
\omega_{i}=max(\frac{cb+b_{0}}{C}-n_{i},0),
\end{equation}
$b$ is the budget per cycle, $b_{0}$ is the size of the initial labeled set, $c \in \{1,2,3,...\}$ denotes the cycle, and $n_i$ is the number of samples selected from class $i$ in previous cycles. Condition \ref{eq:thresh} avoids oversampling from a particular class. To obtain $\Omega$ at cycle $c=1$ for instance, given that we start the AL cycles from uniform initial set with $n_i=b_{0}/C$  we have:
\begin{equation}
\Omega(1) =\frac{b}{C}\mathds{1}_{C\times1}
\end{equation}
In the following, we will minimize Eq.~\ref{eq:balterm} to encourage the selection of class-balanced samples. 

\section{Class Balanced Active Learning}\label{sec:CB}
In this section, we introduce the Class Balanced Active Learning (CBAL) formulation for classification. 

\subsection{Informativeness}\label{informative} 
\paragraph{Entropy} We describe our optimization framework that selects the most uncertain samples while seeking to balance the number of samples over classes.
Based on informativeness approach, given the probability matrix the goal is to find samples that are most uncertain for the model. To measure the uncertainty we use \emph{Entropy}~\cite{dagan1995committee} as an information theory measure that captures the average amount of information contained in the predictive distribution, attaining its maximum value when all classes are equiprobable.  Given the softmax probabilities, the entropy of a sample is computed as:
\vspace{-1mm}
\begin{equation}
\label{eq:probs}
H=-\sum_{\substack{i=1}}^{C} p_{i} \log p_{i}.
\end{equation}
We aim to select samples with maximum entropy. Consequently, the sum of candidates' entropy should also be maximized. In matrix notation form, this is expressed as:
\begin{equation}
\label{eq:entropy}
\sum_{\substack{\{j|z_{j}=1\}}} H(x_{j})=-\boldsymbol{z}^{T}(P\odot \log{(P)}) \mathds{1}_{C\times 1},
\end{equation}
where $\boldsymbol z$ is all-ones column vector and $\odot$ denotes element wise multiplication. $\mathds{1}_{C\times 1}$ is an all-ones column vector. 
In our objective we will minimize the negative entropy, which is equal to maximizing the entropy. 

Finally, we combine the informative and balancing objectives in a single optimization problem given as:
\begin{equation} \label{eq:objective}
\begin{aligned}
\min_{z}\quad
\boldsymbol{z}^{T}(P\odot \log{(P)}) \mathds{1}_{C\times 1}+\lambda\| \Omega(c)-P^T \boldsymbol{z} \|_{1}\\
\textrm{s.t.} \quad \boldsymbol{z}^{T} \mathds{1}_{N\times1}=b, \quad z_i \in \{0,1\}, \quad \forall i =1,2,...,N\\
\end{aligned}
\end{equation}
where $\lambda$ is a parameter that regularizes the contribution of the balancing term in the objective. Minimizing the cost in Eq. \ref{eq:objective} encourages to select sufficient number of samples per class while choosing the most informative ones. The cost function consists of an affine term and a $\ell1$ norm that are both convex, and subsequently their linear combination is also convex. However, as the constraint is non-convex the optimization problem becomes non-convex. The underlying problem is Binary Programming that can be optimally solved by an off-the-shelf optimizer using LP relaxation and the branch and bound method. Algorithm \ref{alg_cyclic} presents the AL cycles using our approach. 

\begin{algorithm}[t]\label{alg_cyclic}
\caption{Class Balancing AL}
\begin{algorithmic}[1]
\renewcommand{\algorithmicrequire}{\textbf{Input:}}
\REQUIRE Unlabeled Pool $\mathcal{D_U}$, Total Budget $B$, Budget Per Cycle $b$,  
\renewcommand{\algorithmicrequire}{\textbf{Initialize:}}
\REQUIRE Initial labeled pool $|\mathcal{D_L}|=b_{0} , c=1$   
\WHILE {$|\mathcal{D_L}| < B$}
    \STATE Train CNN classifier $\Theta$ on $\mathcal{D_L}$
    \STATE Use $\Theta$ to compute probabilities for $x \in \mathcal{D_U}$ 
    \STATE Compute $\Omega(c) $ from Eq. \ref{eq:omega}
    \STATE Solve \ref{eq:objective} or Algorithm 2 for greedy, to obtain $z$
    \STATE Query $z$ to $\mathcal{ORACLE}$ 
    \STATE $\mathcal{D_L} \gets \mathcal{D_L} \cup z, \quad  \mathcal{D_U} \gets \mathcal{D_U} \setminus z$
    \STATE $c \gets c+1$
\ENDWHILE
\RETURN {$\mathcal{D_L}, \Theta$}
\end{algorithmic}
\end{algorithm}
\vspace{-2mm}

\begin{figure*}[ht]
    \centering
    \includegraphics[width=\textwidth]{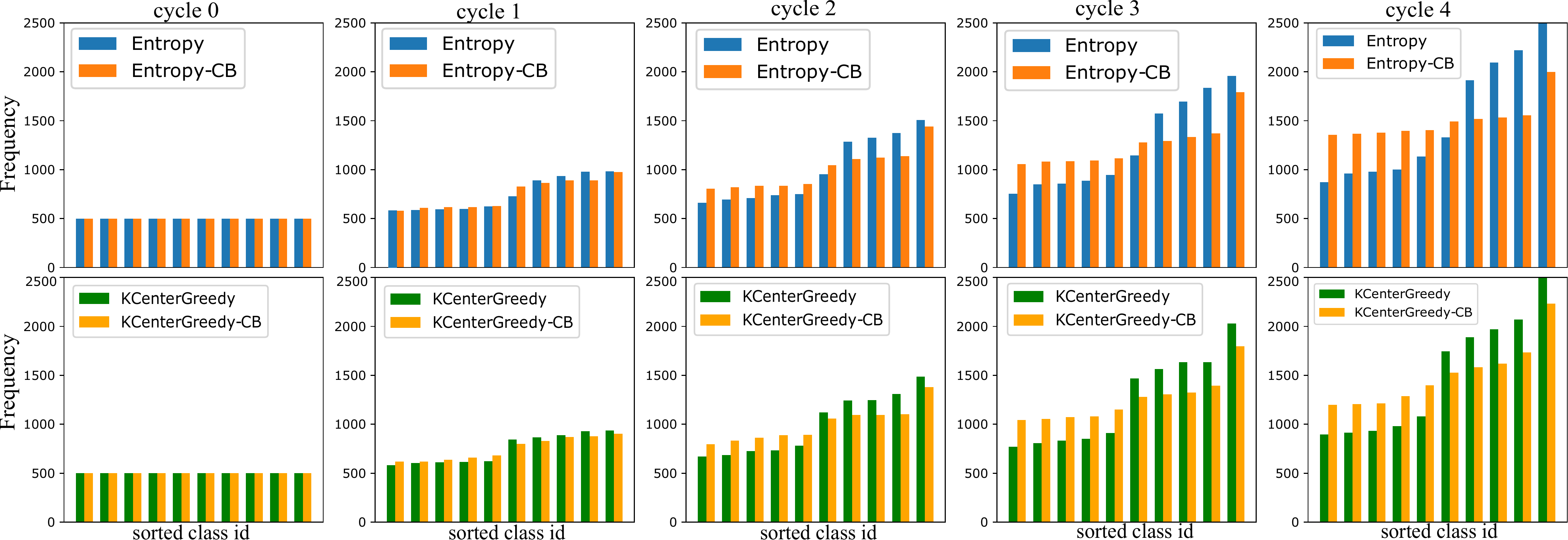}
    \vspace{-4mm}
    \caption{\small \textbf{Class balanced sampling.} Class distribution for Entropy and KCenterGreedy for several active learning cycles on imbalanced CIFAR10 (IF= 0.3.). Our proposed class-balancing (CB) method results in a improved class-balance for both methods.}
    \label{fig:hist2}
    \vspace{-2mm}
\end{figure*}

\begin{figure}[t]
    \centering
    \includegraphics[width=\columnwidth]{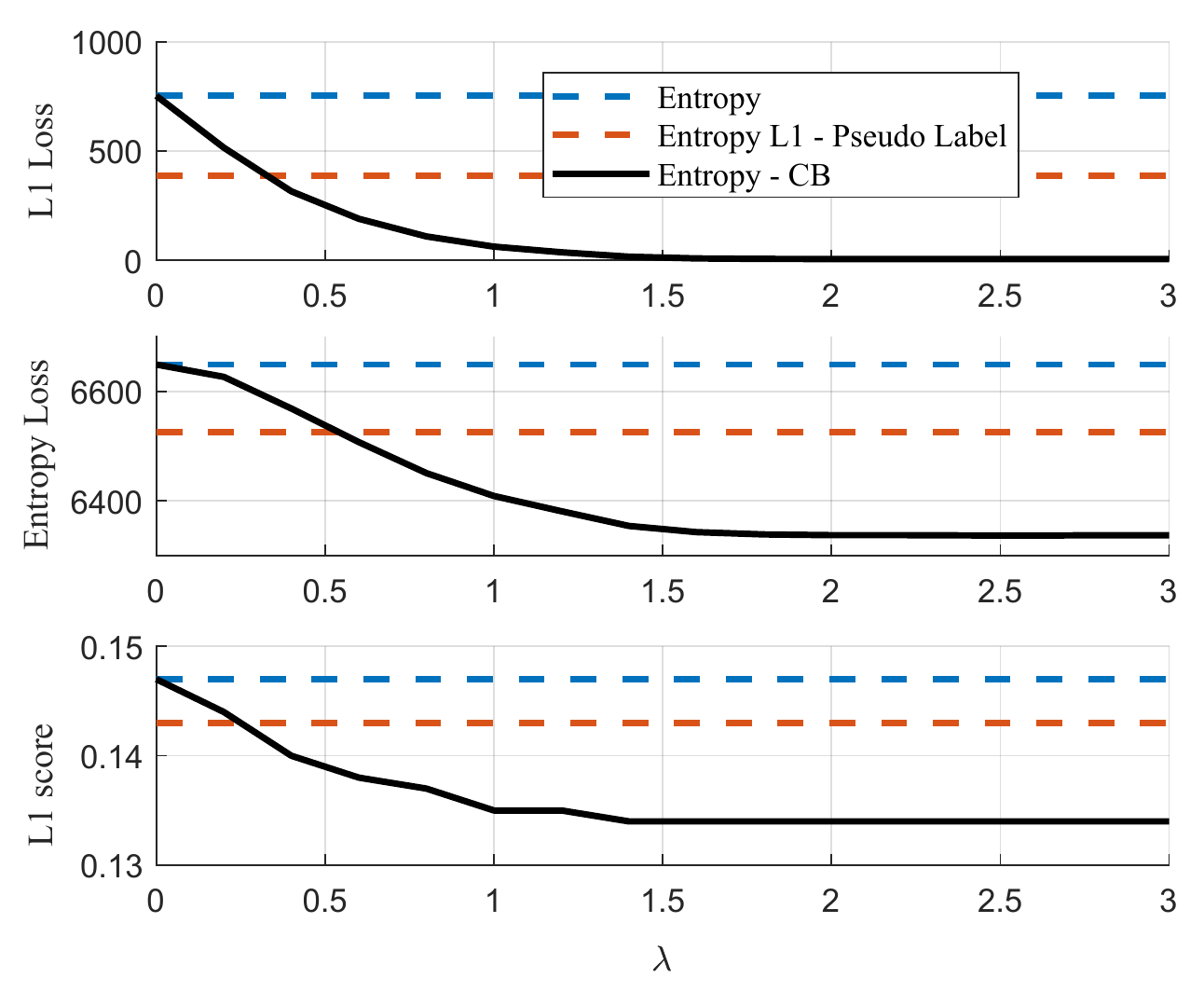}
    \caption{\small \textbf{The effect of $\lambda$ on L1 and entropy losses in the cost function \ref{eq:objective}.}}
    \label{sweepingLambda}
    \vspace{-2mm}
\end{figure}

\paragraph{Regularizer $\lambda$.}
Next, we analyze the effect of varying parameter $\lambda$ on the cost function. We start with a model trained on initial labeled samples of CIFAR100 dataset. Then, for every $\lambda$ in range $(0,3)$ the cost function in Eq.~\ref{eq:objective} is minimized. Fig.~\ref{sweepingLambda} illustrates the changes in entropy loss and the $\ell1$ loss as the components of the cost function with respect to $\lambda$. For comparison purposes, the horizontal lines represent the same losses measured on samples given by standard entropy and entropy L1-pseudo label methods. The latter uses the hard labels given by the model to unlabeled samples also known as "Pseudo Labels" for balancing (see Suppl. Mat. \ref{plot:pseudolabel} for more details and performance evaluation of Entropy-L1-Pseudo Label).
As can be seen, greater $\lambda$ reduces entropy, $\ell1$, and $L1 score$ (introduced in \ref{perf:eval}). It is notable that the samples selected with greater $\lambda$ are more balanced but at the cost of lower entropy. As a result, there is a trade-off between balancedness and entropy of samples.
\vspace{-2mm}
\paragraph{Variational Adversarial Active Learning (VAAL)}
VAAL \cite{Sinha_2019_ICCV} is considered to be one of the current state-of-the-art algorithms on active learning. This model uses a variational autoencoder to map the distribution of labeled and unlabeled data to a latent space. A binary adversarial classifier (analogous to a GAN discriminator) is trained to predict if an image belongs to the labeled or the unlabeled pool. The unlabeled images which the discriminator classifies with lowest certainty as belonging to the labeled pool are considered to be the most representative with respect to other samples which the discriminator thinks belong to the labeled pool. Thus, the images labeled by the discriminator with lower certainty are sampled to be labeled in the next cycle. Considering the uncertainty estimate $u$ of the discriminator, we can encourage finding a balanced sample set by minimizing: 
\vspace{-2mm}
\begin{align}
 \vspace{-2mm}
 \begin{aligned}
 & \qquad\quad \min_{z} \quad \boldsymbol{z}^{T}u +\lambda\| \Omega(c)-P^T \boldsymbol{z} \|_{1}\\
 &\textrm{s.t.} \quad \boldsymbol{z}^{T} \mathds{1}_{N\times1}=b, \quad z_i \in \{0,1\}, \quad \forall i =1,2,...,N\\
 \end{aligned}
 \vspace{-2mm}
\end{align}
\paragraph{Bayesian Active Learning with Disagreement}
BALD method chooses samples that are expected to maximise the information gained about the model parameters. In particular, it select samples that maximise the mutual information between predictions and model posterior \cite{gal2017icml}. It approximates Bayesian inference by drawing Monte Carlo sampling via dropout. Similar to our previous approach, we summarize the mutual information assigned to samples into a vector and incorporate into our optimization problem. 

\subsection{Representativeness}\label{GCBAL}

Representativeness-based methods aim to increase the diversity of the selected batch \cite{sener2018active}. These active learning approaches select the samples iteratively one at a time. In fact, every selected sample influences the next one. Therefore, a method that integrates greedy selection while maintaining the class balance of samples is of great interest. For this reason, we present the greedy class balancing algorithm that incorporates balancing in the sample selection. 

We focus on a prominent method of this approach namely KCenterGreedy, which is a greedy approximation of KCenter problem also known as min-max facility location problem \cite{wolf2011facility}. Our aim is to find $b$ samples having maximum distance from their nearest labeled samples while keeping the samples class-balanced. Similar to \cite{sener2018active}, we compute the embeddings for unlabeled samples via a deep neural network. Specifically, we employ the model for inference on unlabeled samples and consider the penultimate fully connected layer as the visual embedding. Then, we compute the geometrical distances between the representations in the embedding space and construct the distance matrix $D$. Given $N$ unlabeled and $L$ labeled samples, $d_{ij} \in D_{N\times L}$ is the euclidean distance between the embeddings of unlabeled sample $i$ to labeled sample $j$. The algorithm \ref{algGreedy} presents the KCenterGreedy sample selection combined with class balancing. We propose similar cost function to Eq.\ref{eq:objective} for the greedy sampling. In the algorithm $P^{T}z$ represents the cost of already selected samples and matrix $Q$ represents the unlabeled samples to choose from. The broadcasting within the L1 norm is for the consistency across dimensions of labeled samples, unlabeled samples and thresholds. Although here we integrated the balanced sampling with KcenterGreedy, our method is general and applicable to any greedy acquisition method.

\begin{algorithm}[t]\label{algGreedy}
\caption{Greedy Class Balancing Selection}
\begin{algorithmic}[1]
\renewcommand{\algorithmicrequire}{\textbf{Input:}}
\REQUIRE Softmax output $P_{N \times C}$, Distance Matrix $D_{N\times L}$, Balancing threshold $\Omega_{C\times 1}$, Regularizer $\lambda$, Budget Per Cycle $b$ 
\renewcommand{\algorithmicrequire}{\textbf{Initialize:}}
\REQUIRE $z^{(0)}=\mathbf{0}_{N\times1},\quad Q=P$  
\FOR{$i=0:b-1$}
    \vspace{2mm}
    \STATE $d^{(i)}_{N\times1}\gets	min(D,axis=1) \quad \triangleright$ for each unlabeled sample find the nearest labeled sample
    \vspace{2mm}
    \STATE $\psi\gets argmin(-d^{(i)}_{(N-i)\times1} +  \lambda \|\Omega(c)-Q^{T}_{C\times(N-i)}-P^{T}_{C\times N}z^{(i)}\mathds{1}_{1\times(N-i)})\|^{T}_1$
    \vspace{2mm}
    \STATE $z^{(i+1)}(\psi) \gets 1 \qquad \triangleright$ select the sample
    \STATE $Q\gets	P(z^{(i)}=0,:) \quad \triangleright$ keep the remaining unlabeled samples in $Q$
    \vspace{1mm}
    \STATE $D \gets D_{(N-i)\times (L+i)} \quad \triangleright$ update $D$ by removing a row and adding a column correspond to newly selected sample 
\ENDFOR
\RETURN $z^{(b)}$
\end{algorithmic}
\end{algorithm}
\vspace{-2mm}
\section{Experiments}\label{sec:exp}
\subsection{Experimental Setup}
\vspace{-2mm}
We evaluate our method on three image classification benchmarks and the imbalanced variants. The initial labeled set $\mathcal{D_L}$ consists of 10$\%$ of the training dataset that is uniformly selected from all classes at random. At each cycle we start with our base model either from scratch or, in case of Tiny-imagenet, we start from a pretrained imagenet model. We train the model in $c$ cycles until the budget $B$ is exhausted. The budget per cycle for all experiments is 5$\%$ of the original dataset.

\minisection{Datasets.}
To evaluate our method, we use CIFAR10 and CIFAR100 \cite{krizhevsky2012learning} datasets with 50K images for training and 10K for test. CIFAR10 and CIFAR100 have 10 and 100 object categories respectively and an image size of 32$\times$32.
To evaluate the scalability of our method we evaluate on Tiny ImageNet dataset \cite{le2015tiny} with 90K images for training and 10K for testing. There are 200 object categories in Tiny ImageNet with an image size of 64$\times$64. 
\vspace{-1mm}
\begin{figure*}[t]
    \centering
    \includegraphics[width=\textwidth]{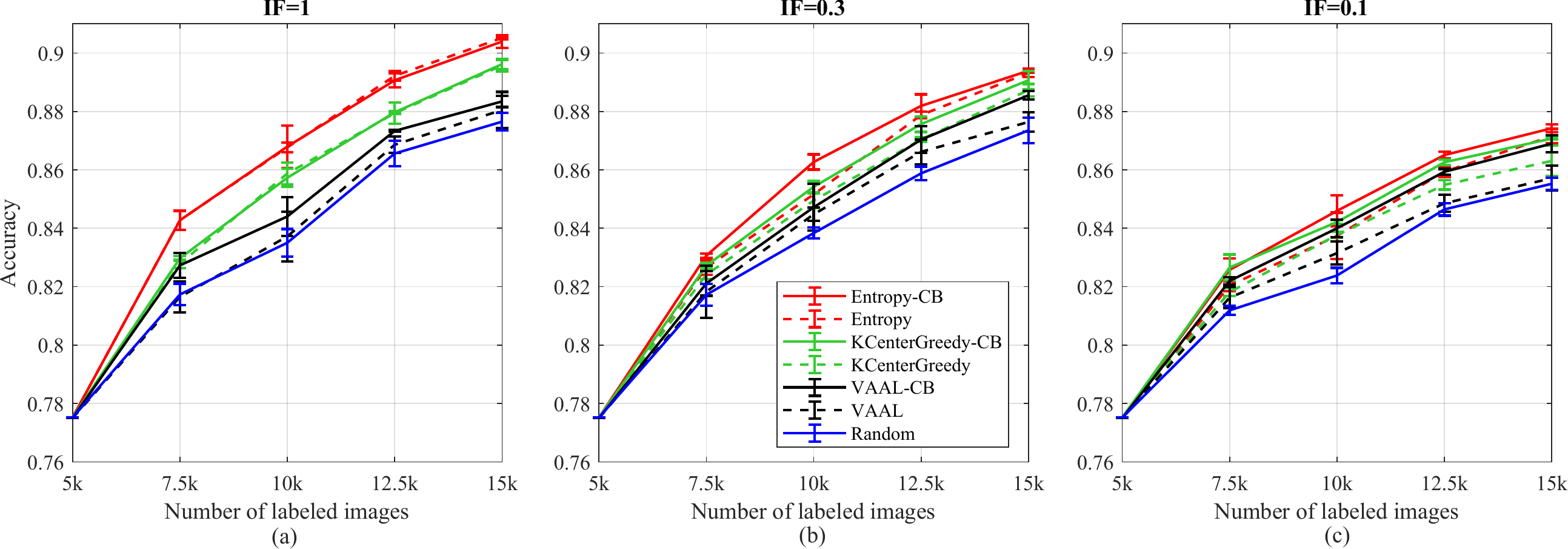}
    \caption{\small \textbf{Performance evaluation.} Results for several active learning methods on CIFAR10 for different imbalance factors (IF).}
    \label{fig:cifar10}
    \vspace*{-5mm}
\end{figure*}
\vspace*{-4mm}
\paragraph{Long-Tailed Datasets.}
To verify our approach on imbalanced datasets, we make the CIFAR10, CIFAR100 and Tiny ImageNet class-imbalanced. Again, we reserve 10$\%$ of samples of the three datsasets for initial labeled set. As in \cite{cui2019class} we create long-tailed datasets with the remaining 90$\%$ by randomly removing training examples. In particular, the number of samples drops from $y$-th class is $n_{y}\cdot \textrm{IF}$ where $n_{y}$ is the original number of training samples in class $y$ and the imbalance factor $\textrm{IF} \in (0,1)$. For the construction of long-tailed datasets we apply IF to half of the classes, and use $\textrm{IF}\in\{0.1, 0.3\}$.
\vspace*{-5mm}
\paragraph{Baselines.}
We compare our method with Random sampling and several informative and representative-based approaches including  Entropy sampling, KCenterGreedy, VAAL and BALD (See also suppl. mat. \ref{plot:coreset} for Coreset ). In order to make a fair comparison with the baselines, we used their official code and adapted them into our code to ensure an identical setting.
\vspace{-5mm}
\paragraph{Performance Evaluation.} \label{perf:eval}
We measure the accuracy on the test set to evaluate the performance. Results for all experiments are averaged over 3 runs. For each method we plot the average performance for all runs with vertical bars to represent the standard deviation.
To measure the balancedness of selected samples, we use $L1\_score$  by computing $\ell1$ distance between samples' distribution and uniform distribution. In order to have a measure ranging from 0 to 1, we normalize $\ell1$ with the factor obtained as follows:
\begin{equation}
\begin{split}
&\ell1([b,0,,..,0], [\frac{b}{C},...,\frac{b}{C}])= \\
& |b-\frac{b}{C}|+|0-\frac{b}{C}|+...+|0-\frac{b}{C}|=\frac{2b(C-1)}{C}.
\end{split}
\end{equation}
The first argument represents the distribution in which the entire budget $b$ is spent to sample from a single class while the second argument represents the uniform sampling.

\begin{figure*}[t]
    \centering
    \includegraphics[width=\textwidth]{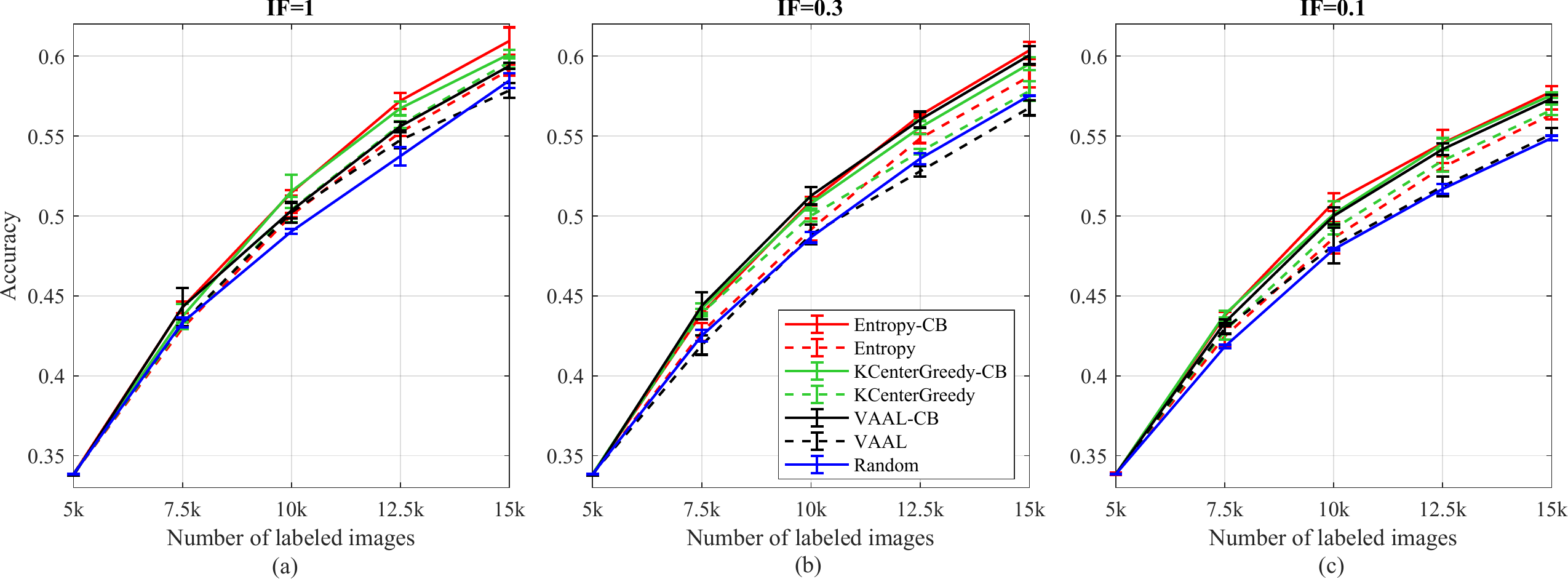}
    \caption{\small \textbf{Performance evaluation.} Results for several active learning methods on CIFAR100 for different imbalance factors (IF).}
    \label{fig:cifar100}
    \vspace*{-4mm}
\end{figure*}
\vspace{-4mm}
\paragraph{Implementation details.}
Our method is implemented in PyTorch \cite{paszke2017automatic}. We start with Resnet18 \cite{He_2016} trained from scratch every cycle. For Tiny-Imagenet dataset however we start with pretrained ImageNet model and Resnet101. All the models are trained with SGD optimizer with momentum 0.9 and an initial learning rate of 0.02 and 0.01 for CIFAR10/100 and Tiny ImageNet respectively. We train CIFAR datasets for 100 epochs and reduce the learning rate by a factor of 0.5 once at 60 and again at 80 epochs. In the case of Tiny ImageNet we reduce the learning rate at 10, 15, 20, 25 epochs by factor of 0.5 training for a total of 30 epochs. 
During training, we apply a standard augmentation
scheme including random crop from zero-padded images, random horizontal flip, and image normalization using the channel mean and standard deviation estimated over the training set. 
We set the regularizer $\lambda$ based on the analysis in \ref{sweepingLambda} specifically for each method. We use the model trained on initial labeled pool to set lambda. We choose the smallest $\lambda$ after which the L1 loss converges and does not diminish further. 
Once we chose $\lambda$ we keep it fixed for that method across the experiments.
To efficiently solve the optimization problem we used python CVXPY \cite{diamond2016cvxpy} with Gurobi solver \cite{gurobi}.
\subsection{Experimental Results}
\paragraph{Performance on CIFAR10.}
Fig.~\ref{fig:hist2} provides an evaluation of the class balancing technique on Entropy and KcenterGreedy. The distribution of samples selected by Class Balanced (CB) methods evidently remains close to the uniform compared to the baselines across cycles. Fig.~\ref{fig:cifar10} presents the quantitative results on CIFAR10. Dashed curves represent the standard methods and solid curves represent those equipped with class-balancing. We start by evaluating the performance on the balanced (original) dataset denoted by IF=1. We observe in Fig.~\ref{fig:cifar10}.a that the addition of class balancing gives similar results compared to the standard methods. However, for the case of VAAL, class-balancing results in notable improvements. Next, we evaluate the performance of class-balancing on the imbalanced CIFAR10 dataset where IF=0.3. Fig.~\ref{fig:cifar10}.b illustrates clearly how class-balancing is beneficial for all methods across the cycles. The class-balanced variants constantly improve the performance of both informative and representative baselines. Regarding the active learning gain, Entropy-CB achieves the performance of $86\%$ whereas Random requires almost $10\%$ more annotation (equivalent to 5K images) to achieve the same performance.
Fig.~\ref{fig:cifar10}.c illustrates the performance on a severely imbalanced dataset where IF=0.1. We observe a considerable improvement using class balancing over the baselines. In particular VAAL-CB achieves a growing improvement of $1\%$ on average over VAAL across the cycles. See table~\ref{tab:CIFA10} for analysis of performance gains over baselines.
\vspace{-4mm}
\paragraph{Performance on CIFAR100.}
Fig.~\ref{fig:cifar100} presents the performance 
on CIFAR100. In Fig.~\ref{fig:cifar100}.a, the class balanced methods improve baselines marginally even though the dataset is balanced (IF=1). The improvements of class-balanced methods improve for the lower IF values (see Fig~\ref{fig:cifar100}.b and c).
Notably, VAAL-CB achieves $3\%$ improvement on average over the VAAL baseline in Fig.~\ref{fig:cifar100}.b. See Table~\ref{tab:CIFAR100} for a detailed gain analysis. To put these improvements into perspective, the gain obtained by class balancing methods over the baselines is comparable to the improvement of those methods over Random. Specifically, Entropy-CB after 4 cycles achieves over $1\%$ improvement over the Entropy baseline regardless of imbalance factor of the dataset.

\begin{table}[t]
    \centering
    \resizebox{\columnwidth}{!}{
    \begin{tabular}{ll ccccc}
        \toprule
        \midrule
         \multirow{2}{*}{\bf Imbalance Factor} & \multirow{2}{*}{\bf Methods}&
         &\multicolumn{2}{c}{\bf Cycles} \\
         &  & 1 & 2 & 3 & 4 \\
        \midrule
        \multirow{3}{*}{\bf IF=0.1} & Entropy CB(\%) & 0.54 &   0.86  &  0.57  &  0.27 \\ 
             & KcenterGreedy CB(\%) & 0.84  &  0.44  &  0.77  &  0.77  \\
             & VAAL CB(\%) & 0.57  & 0.85 &  1.08  &  1.19\\
        \midrule
        \multirow{3}{*}{\bf IF=0.3} & Entropy CB(\%) & 0.40  &  1.11 &   0.31  &  0.08 \\
             & KcenterGreedy CB(\%) & 0.31  &  0.47  &  0.53  &  0.34 \\
             & VAAL CB(\%) & 0.28 &   0.24 &   0.42  &  0.91 \\
        \midrule
        \multirow{3}{*}{\bf IF=1} & Entropy CB(\%) &  0.00  &  0.027 &  -0.15  & -0.12 \\
             & KcenterGreedy CB(\%) & 0.19 &  0.14 &   0.03  &  0.05   \\
             & VAAL CB(\%) & 1.08  &  0.68 &   0.47 &   0.29 \\     \midrule        
        \bottomrule
    \end{tabular}}
    \vspace{2mm}
    \caption{\small  \textbf{Performance gain over AL baselines on CIFAR 10.}}
    \label{tab:CIFA10}
    \vspace*{-2mm}
\end{table}

\begin{table}[t]
    \centering
    \resizebox{\columnwidth}{!}{
    \begin{tabular}{ll ccccc}
        \toprule
        \midrule
         \multirow{2}{*}{\bf Imbalance Factor} & \multirow{2}{*}{\bf Methods}&
         &\multicolumn{2}{c}{\bf Cycles} \\
         &  & 1 & 2 & 3 & 4 \\
        \midrule
        \multirow{3}{*}{\bf IF=0.1} & Entropy CB(\%) & 1.28 &  2.23  &  1.50  &  1.43\\ 
             & KcenterGreedy CB(\%) & 1.03  &  0.92  &  1.04  &  0.93 \\
             & VAAL CB(\%) & 0.37 &   1.86  &  2.32  &  2.23 \\
        \midrule
        \multirow{3}{*}{\bf IF=0.3} & Entropy CB(\%) & 1.16 & 1.76  &  1.44 & 1.63 \\
             & KcenterGreedy CB(\%) & 0.28  &  0.76  &  1.52  &  1.70   \\
             & VAAL CB(\%) & 2.47 &  2.42  &  3.23 &  3.29 \\
        \midrule
        \multirow{3}{*}{\bf IF=1} & Entropy CB(\%) & 1.37  &  1.40  &  1.96  &  1.82   \\
             & KcenterGreedy CB(\%) & 0.55  &  1.15  &  1.03  &  0.48  \\
             & VAAL CB(\%) & 1.01  &  0.11  &  0.86  &  1.53  \\ \midrule             
        \bottomrule
    \end{tabular}}
    \vspace{2mm}
    \caption{\small  \textbf{Performance gain over AL baselines on CIFAR 100.}}
    \label{tab:CIFAR100}
    \vspace*{-4mm}
\end{table}

\begin{table}[t]
    \centering
    \resizebox{\columnwidth}{!}{
    \begin{tabular}{ll ccccc}
        \toprule
        \midrule             
         \multirow{2}{*}{\bf Imbalance Factor} & \multirow{2}{*}{\bf Methods}&
         &\multicolumn{2}{c}{\bf Cycles} \\
         &  & 1 & 2 & 3 & 4 \\
        \midrule
        \multirow{2}{*}{\bf IF=0.1} & Entropy CB (\%) & 0.48 & 0.63 & 0.21 & 0.58 \\ 
             & BALD CB(\%) & 0.31 & 0.10 &0.08 &0.21 \\
        \midrule
        \multirow{2}{*}{\bf IF=0.3} & Entropy CB (\%) & 0.34 & -0.04 & 0.52 & 0.19\\
             & BALD CB(\%) & 0.07 & 0.07 & 0.36& 0.19\\
        \midrule
        \multirow{2}{*}{\bf IF=1} & Entropy CB(\%) & 0.35 & 0.62 & 0.56 & 0.74  \\
             & BALD CB(\%) & 0.10& 0.21 &1.11 & 0.51  \\
        \midrule             
        \bottomrule
    \end{tabular}}
    \vspace{2mm}
    \caption{\small  \textbf{Performance gain over baselines on Tiny ImageNet.}}
    \label{tab:tinyImg}
    \vspace*{-4mm}
\end{table}
\vspace*{-3mm}
\paragraph{Performance on Tiny ImageNet.}
Tiny ImageNet is a challenging large scale dataset which we use to evaluate the scalability of our approach. Also to evaluate the generality of our approach we show the performance of class balancing applied to BALD as a Baysian approach\footnote{Representativeness-based methods are infeasible on large datasets.}. Table\ref{tab:tinyImg}  shows evidently the addition of class balancing to Entropy and BALD boost their performance on both balanced and imbalance datasets. (See suppl. mat. \ref{sec:tinyImg} for a detailed performance evaluation on Tiny ImageNet).    

\vspace*{-2mm}
\section{Conclusions}
\vspace*{-2mm}
We have investigated the influence of class-imbalance on active learning performance. Class-imbalance can be caused by an imbalanced unlabeled data pool or by the sampling bias present in active learning algorithms. When aiming for good performance of the final classifier on all classes, class-imbalance has a detrimental effect. Therefore, to address  class-imbalance we proposed an optimization-based method that aims to balance classes. The method is general and can be combined with both the informativeness and representativeness criteria often used in active learning. Extensive experiments, on several  datasets show that our method improves results of existing active learning methods. Our results suggests that class-balancing should be an important criteria when selecting samples, and that it should be considered next to the long-standing active learning criteria of informativeness and representativeness. 

\textbf{Acknowledgements} We ackowledge the support of the project  PID2019-104174GB-I00 (MINECO, Spain), the CERCA Programme of Generalitat de Catalunya, the EU project CybSpeed MSCA-RISE-2017-777720 and CYTED Network (Ref. 518RT0559).

{\small
\bibliographystyle{ieee_fullname}
\bibliography{egbib}
}

\renewcommand\thesection{\Alph{section}}
\renewcommand\thesubsection{\thesection.\Alph{subsection}}

\onecolumn
\begin{center}
\textbf{{\LARGE Supplementary Materials for \\
Class-Balanced Active Learning for Image Classification}}
\newline
\end{center}

\section{Performance on Tiny ImageNet dataset}\label{sec:tinyImg}
\begin{figure}[th]
    \centering
    \includegraphics[width=\textwidth]{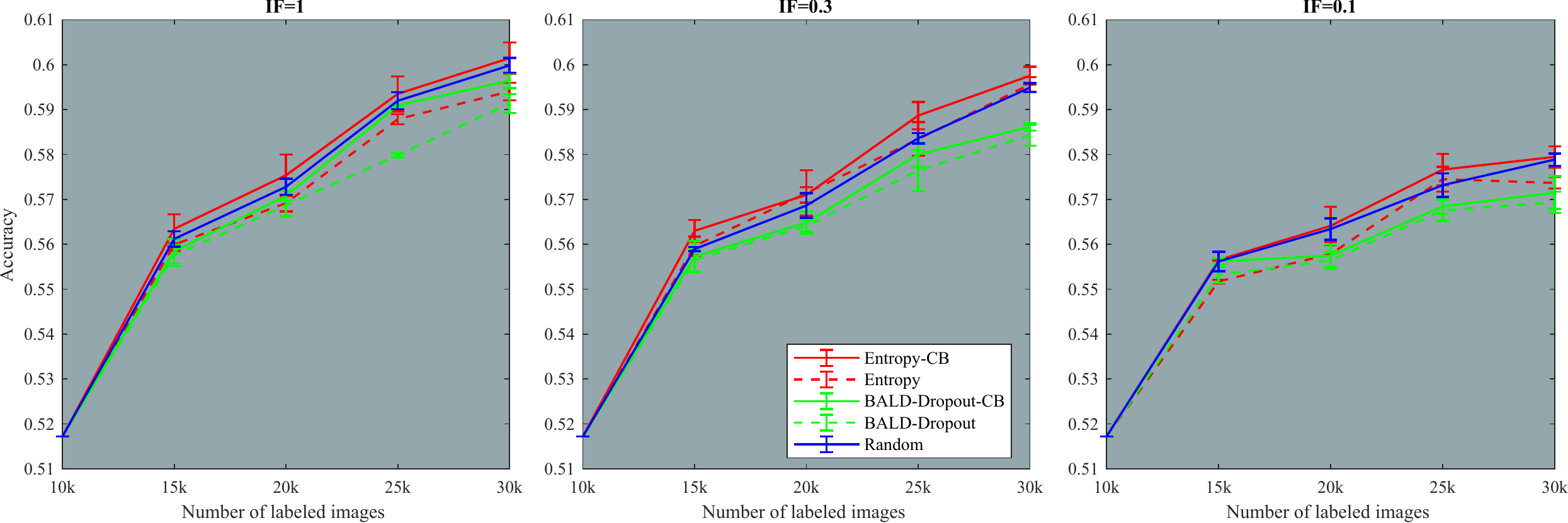}
    \caption[width=2\textwidth]{\small \textbf{Performance evaluation.} Results for active learning methods on Tiny ImageNet with different imbalance factors (IF).}
    \label{figtinyImgNet}
\end{figure}

Fig. \ref{figtinyImgNet} illustrates the performance of class balanced (CB) methods and AL baselines. As can be seen, both Entropy-CB and BALD-CB outperform the corresponding baselines. Notably in Tiny ImageNet, Random sampling serve as a competitive baseline. Nevertheless the addition of class balancing made Entropy-CB superior in almost all active learning cycles across different imbalance factors.      
\section{Pseudo Label balancing}\label{plot:pseudolabel}
\begin{figure}[p]
    \centering
    \includegraphics[width=\textwidth]{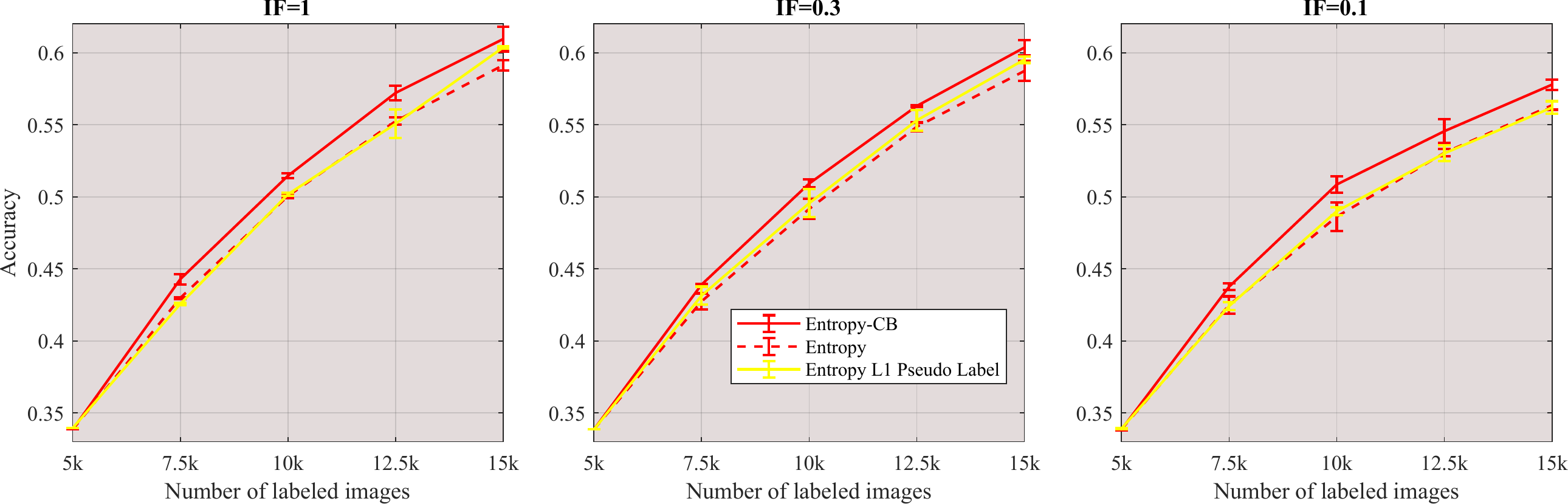}
    \caption{\small \textbf{Performance evaluation.} Comparing Entropy standard, Entropy balanced by Pseudo Labels against the proposed Entropy CB.}
    \label{fig:pseudo}
\end{figure}

Fig.~\ref{fig:pseudo} presents the performance of another Entropy variation  on CIFAR100 for comparison. Among them, "Entropy L1 Pseudo Label"  benefits from "pseudo labels" defined as the most probable labels that the model assigns to unlabeled samples (the prediction of the model is then converted to a one-hot vector).  This method utilizes the pseudo labels to balance the distribution of samples and select certain number of samples (specified by $\Omega$ in Eq.3) from each class with maximum entropy. The experiments show that Entropy-CB outperforms Entropy L1 Pseudo Label both in terms of active learning performance (see Fig.~\ref{fig:pseudo}) and the ability of class balancing (see $\lambda$ tuning in Section 4).   

\section{CoreSet performance}\label{plot:coreset}
\begin{figure}[th]
    \centering
    \includegraphics[width=\textwidth]{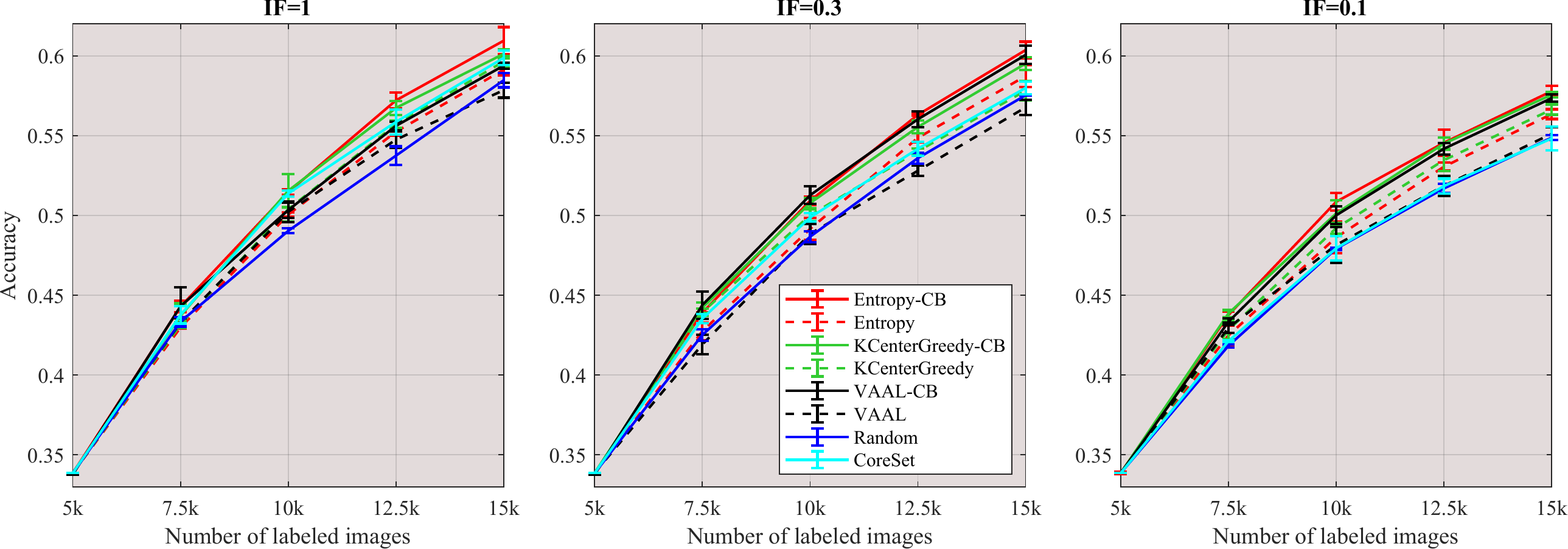}
    \caption{\small \textbf{Performance evaluation.} CoreSet compared to active learning methods on CIFAR100 with different imbalance factors (IF).}
    \label{fig:coreset_cifar100}
\end{figure}

The performance of CoreSet on CIFAR10 and CIFAR100 is shown in Fig.~\ref{fig:coreset_cifar10} and Fig.~\ref{fig:coreset_cifar100} resepectively. In our experiments CoreSet and KCenterGreedy-CB perform similarly on the balanced dataset (IF=1). However, when the dataset is imbalanced (IF=0.3 and IF=0.1) the performance of CoreSet degrades compared to KCenterGreedy-CB. As CoreSet is a MIP (Mixed Integer Programming) problem, 
our technique cannot be applied to this method. 
\begin{figure}[th]
    \centering
    \includegraphics[width=\textwidth]{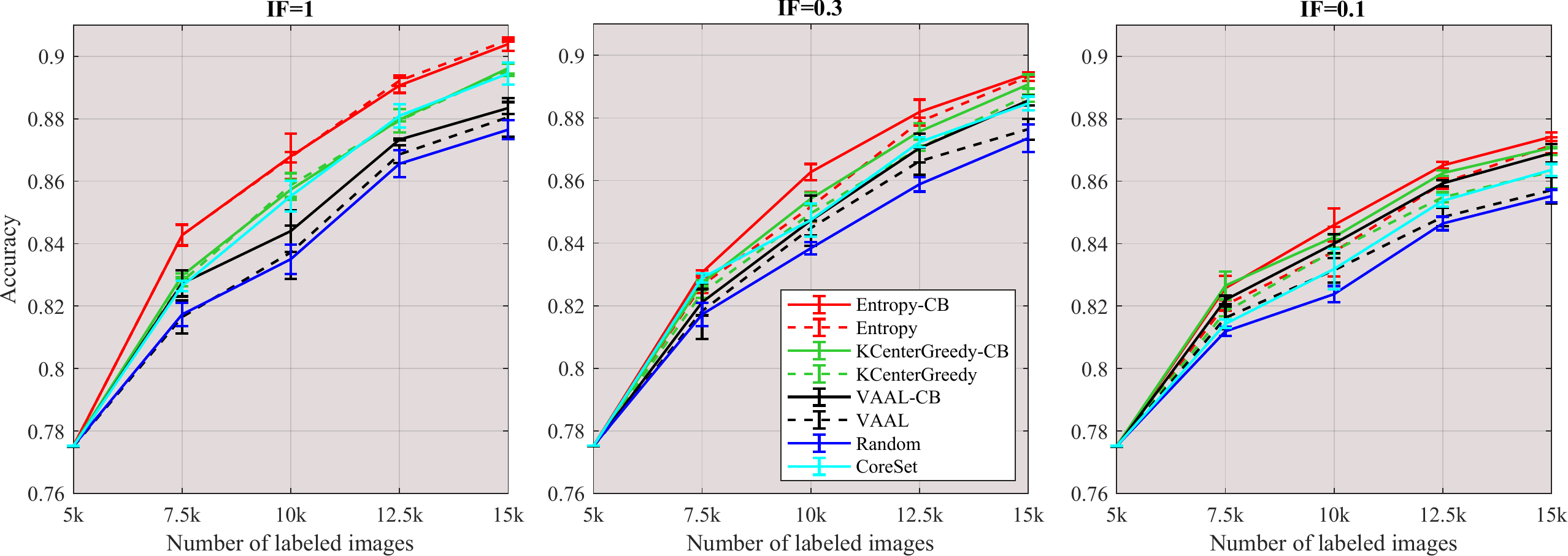}
    \caption{\small \textbf{Performance evaluation.} CoreSet compared to active learning methods on CIFAR10 with different imbalance factors (IF).}
\end{figure}\label{fig:coreset_cifar10}

\clearpage

\section{Distribution of selected samples in CIFAR100}\label{hist_5_cycles:cifa100}
\begin{figure*}[h]
    \centering
    \includegraphics[width=\textwidth]{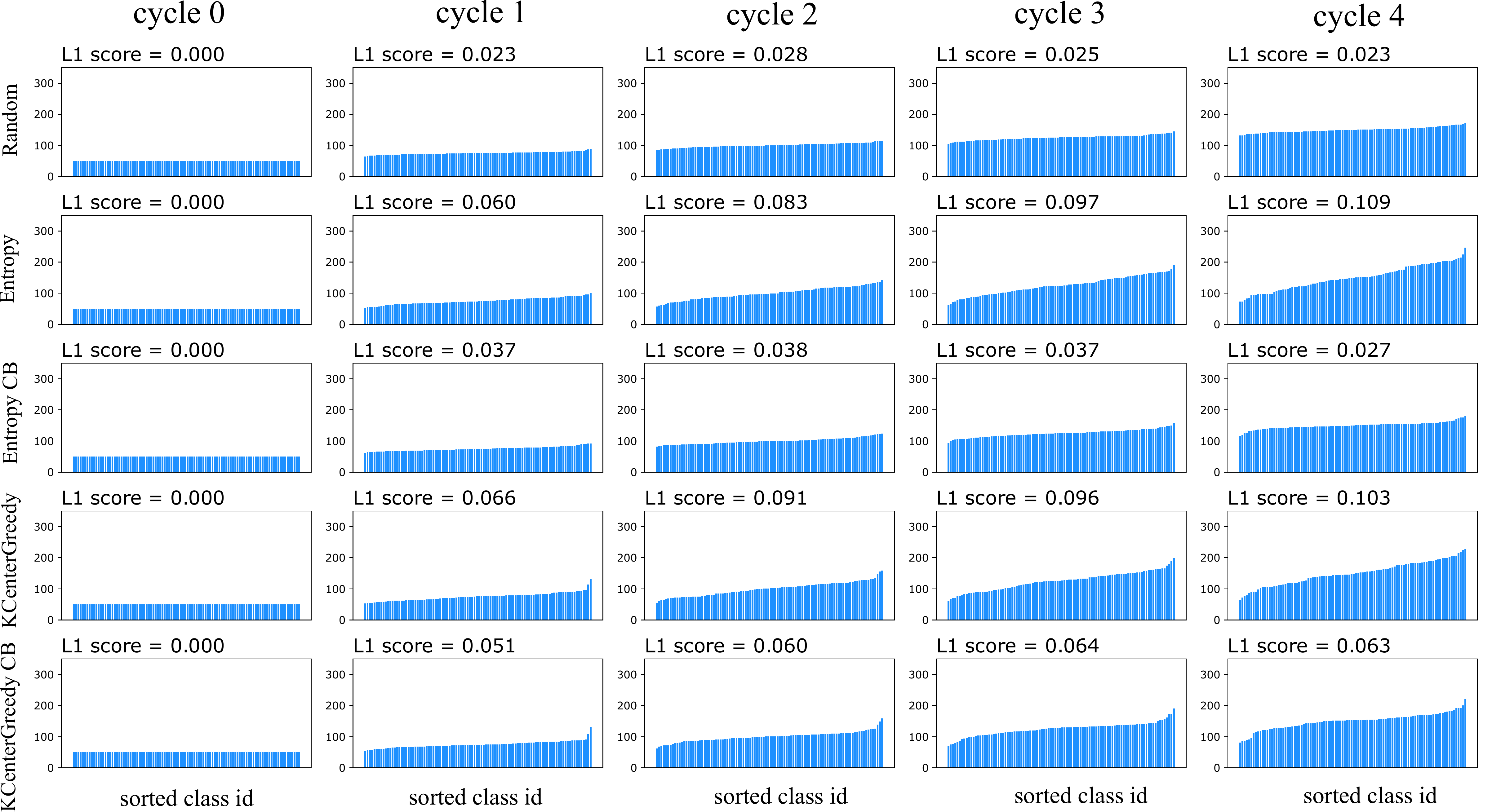}
    \caption{\small \textbf{Distribution of samples selected by our proposed method (CB) compared to baselines on CIFAR100 with IF=1.}}
    \label{fig:class_dist_1}
\end{figure*}

\begin{figure}[t]
    \centering
    \includegraphics[width=\textwidth]{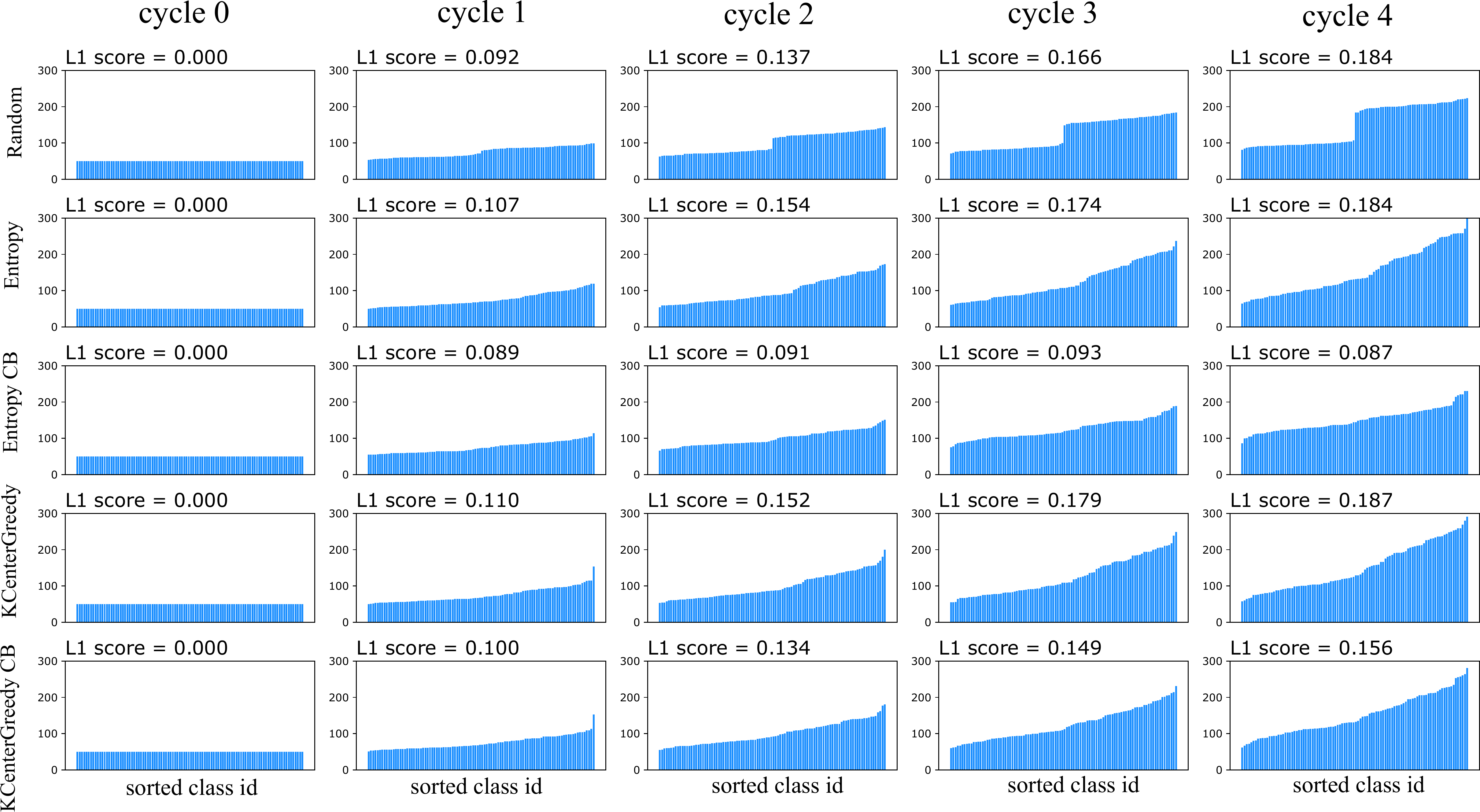}
    \caption{\small \textbf{Distribution of samples selected by our proposed method (CB) compared to baselines on CIFAR100 with IF=0.3.}}
    \label{fig:class_dist_03}
\end{figure}

\begin{figure}[t]
    \centering
    \includegraphics[width=\textwidth]{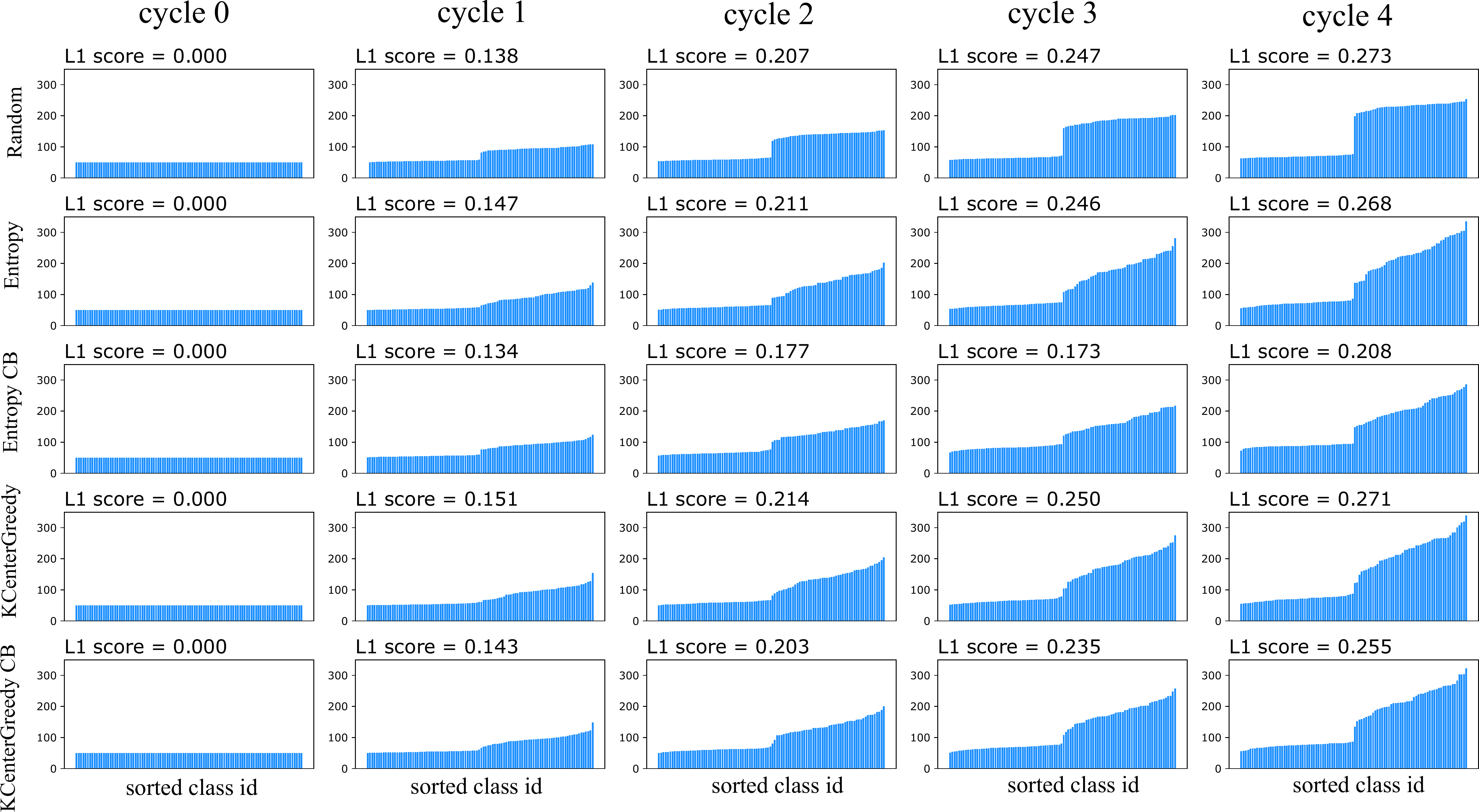}
    \caption{\small \textbf{Distribution of samples selected by our proposed method (CB) compared to baselines on CIFAR100 with IF=0.1.}}
    \label{fig:class_dist_01}
\end{figure}

\vspace{5mm}
Fig.~\ref{fig:class_dist_1}, ~\ref{fig:class_dist_03} and ~\ref{fig:class_dist_01} show the distribution of samples selected by AL methods on original (IF=1), imbalanced (IF=0.3) and (IF=0.1) respectively. The L1 score above the distributions (introduced in Section 5.1)  measures the $\ell1$ distance from uniform distribution in the corresponding cycle. As can be seen, CB methods are remarkably effective in balancing the distribution of selected samples regardless of imbalance factor. It is worth mentioning in Fig ~\ref{fig:class_dist_1} although the dataset is balanced, AL baselines (Entropy and KCenterGreedy) result in biased sampling. In contrast, CB methods provide more balanced samples across all cycles and imbalance factors.

\vspace{5mm}

\end{document}